Learning Under Extreme Data Scarcity: Subject-Level Evaluation of Lightweight CNNs for fMRI-Based Prodromal Parkinson's Detection

Naimur Rahman
Bath Spa University, Bath, United Kingdom
Contact: **n.rahman@bathspa.ac.uk**

**Abstract**

Deep learning is often applied in settings where data are limited, correlated, and difficult to obtain, yet evaluation practices do not always reflect these constraints. Neuroimaging for prodromal Parkinsons disease is one such case, where subject numbers are small and individual scans produce many highly related samples. Here, we examine prodromal Parkinsons detection from resting state fMRI as a machine learning problem centered on learning under extreme data scarcity.

Using fMRI data from 40 subjects, consisting of 20 prodromal Parkinsons cases and 20 healthy controls, ImageNet pretrained convolutional neural networks are fine-tuned and evaluated under two different data partitioning strategies. Results show that commonly used image level splits allow slices from the same subject to appear in both training and test sets, leading to severe information leakage and near perfect accuracy. When a strict subject level split is enforced, performance drops substantially, yielding test accuracies between 60 and 81 percent.

Models with different capacity profiles are compared, including VGG19, Inception V3, Inception ResNet V2, and the lightweight MobileNet V1. Across subject-level evaluations, MobileNet shows the most reliable generalization, outperforming deeper architectures despite having significantly fewer parameters. Taken together, the results indicate that in low-data regimes, evaluation strategy and model capacity have a greater impact on performance than architectural depth. We further quantify how naive image-level splitting can inflate apparent accuracy to nearly 100% through information leakage, whereas subject-level splits yield substantially lower but more realistic performance estimates (60–81% accuracy). Although our analysis is limited to 40 subjects from a single cohort and does not include external validation or cross-validation, it provides a concrete case study and practical recommendations for evaluating deep models under extreme data scarcity.

Taken together, the results indicate that in low data regimes, evaluation strategy and model capacity have a greater impact on performance than architectural depth. Lightweight networks can generalize more reliably than complex models when data are scarce, while improper validation can lead to misleading conclusions. Although demonstrated on prodromal Parkinsons detection, the findings extend to other data limited and safety critical machine learning applications where robustness and evaluation integrity are essential.

## 1. Introduction

Machine learning models are increasingly applied in settings where labelled data are limited, highly correlated, and expensive to obtain. Medical neuroimaging is a clear example, particularly for prodromal Parkinsons disease, where confirmed cases are rare and datasets are necessarily small. Parkinsons disease is a neurodegenerative disorder in which motor symptoms typically emerge only after substantial neuronal loss, while an earlier prodromal phase may present through subtle non motor manifestations such as REM sleep behaviour disorder or olfactory dysfunction (Chaudhuri et al., 2006; Postuma and Berg, 2016). Identifying individuals during this phase is of high clinical interest, yet reliable biomarkers remain difficult to establish, and available datasets are often extremely constrained in size.

Rather than approaching this problem primarily from a clinical perspective, this work treats prodromal Parkinsons detection as a methodological case study in learning under extreme data scarcity. Using resting state fMRI data from only 40 subjects, consisting of 20 prodromal Parkinsons cases and 20 healthy controls from the Parkinsons Progression Markers Initiative cohort (Jennings et al., 2014), we investigate how deep learning models behave when the number of independent training samples is very small. The focus is on how evaluation strategy and model capacity influence apparent performance, and how misleading conclusions can arise if these factors are not handled carefully.

A central challenge in small scale medical imaging studies is information leakage during model evaluation. When multiple samples derived from the same subject, such as two-dimensional MRI slices, appear in both training and test sets, a model may exploit subject specific characteristics rather than learning disease related patterns. This can result in deceptively high accuracy that does not reflect true generalization. Several prior studies have reported near perfect classification performance for Parkinsons disease using convolutional neural networks applied to MRI data (Shinde et al., 2019; Mostafa and Cheng, 2020), but such results are increasingly understood to be sensitive to data partitioning strategies. Analyses of neuroimaging pipelines have shown that slice level evaluation can substantially inflate reported accuracy compared to subject level evaluation, even in cases where class labels are randomized, underscoring the severity of this issue (Postuma and Berg, 2016). Recent work in medical machine learning therefore emphasizes strict subject wise splits and careful validation as essential requirements for trustworthy evaluation.

Model complexity presents an additional challenge in low data regimes. Deep architectures with large numbers of parameters can easily overfit when training data are limited, raising the question of whether model capacity should be reduced rather than increased in such settings. While larger networks offer greater representational power, they often require more data to generalize reliably. Conversely, compact architectures may provide implicit regularization and improved stability when information is scarce. Prior work has suggested that lightweight convolutional networks can outperform deeper models on small medical imaging datasets (Howard et al., 2017; Amoroso et al., 2018). Motivated by this observation, we compare architectures spanning a wide range of capacities, from very deep models such as Inception

ResNet V2 to lightweight models such as MobileNet V1, under identical training and evaluation conditions.

The contributions of this work are threefold. First, we provide a concrete empirical demonstration of how naive slice-level data splitting can lead to misleadingly high accuracy in small-sample neuroimaging, while strict subject-level evaluation yields far more conservative and plausible performance estimates. Second, we analyze the relationship between model capacity and generalization in an extreme low-sample setting, showing that a lightweight architecture achieves stronger and more stable performance than deeper alternatives despite having substantially fewer parameters. Third, we distil practical recommendations for model selection and evaluation on small, hierarchically structured datasets, highlighting the importance of subject-wise partitioning, capacity-aware architecture choice, and transparent reporting of evaluation protocols. Although demonstrated on prodromal Parkinson's fMRI data, these methodological lessons extend to other machine learning applications where data scarcity and evaluation integrity are critical concerns.

## 2. Related Work
### 2.1 Small-sample neuroimaging and PD detection

Machine learning applied to neuroimaging frequently operates under severe data constraints. In disorders such as Parkinsons and Alzheimer's disease, early studies reported high classification accuracy using convolutional neural networks on MRI data, but subsequent analyses have shown that many of these results were overly optimistic due to methodological biases. Reported accuracies in the range of 88 to 100 percent for Parkinsons disease classification have since been recognized as unrealistic for true generalization, particularly when subject counts are low (Shinde et al., 2019; Mostafa and Cheng, 2020).

A primary source of inflated performance in these studies is data leakage introduced during evaluation. Common issues include slice level data partitioning, where multiple samples derived from the same subject appear in both training and test sets, as well as hyperparameter tuning performed on test data. Such practices allow models to exploit subject specific characteristics rather than disease related patterns, leading to misleading validation scores. As a result, recent work in medical machine learning has emphasized the importance of strict subject wise data splits and rigorous validation protocols, especially in small datasets (Postuma and Berg, 2016).

Our work is situated within this context. We use prodromal Parkinsons disease as a case study to quantify the impact of slice level leakage and to demonstrate the necessity of subject level evaluation. Prodromal Parkinsons detection remains an open and challenging problem, and prior efforts have explored a wide range of potential biomarkers, including neuroimaging, olfactory testing, sleep behavior analysis, and wearable sensors (Iakovakis et al., 2020; Tracy et al., 2020). In this study, we focus on resting state fMRI as a high dimensional but promising modality, while treating the task primarily as a methodological investigation rather than a clinical diagnostic solution.

### 2.2 Deep CNNs and transfer learning in neuroimaging

Convolutional neural networks are now widely used for image analysis and have been applied to MRI based disease classification with mixed results. Given the limited size of most medical imaging datasets, transfer learning from large scale computer vision datasets such as ImageNet has become a standard strategy. Pretrained convolutional networks provide generic feature representations that can be adapted to medical images with relatively few labelled samples (Simonyan and Zisserman, 2014; Szegedy et al., 2016).

In the context of Parkinsons disease, CNN based approaches have been explored using structural MRI, neuromelanin sensitive MRI, and DaT scans, with varying degrees of success (Shinde et al., 2019; Mostafa and Cheng, 2020). While some studies report strong performance, such results must be interpreted cautiously, as high model capacity combined with limited subject numbers increases the risk of overfitting. Ensemble methods can further boost apparent accuracy in some cases, but they also introduce additional complexity and may amplify the tendency to memorize subject specific artifacts when data are scarce (Mostafa and Cheng, 2020). When subject counts are low, models may inadvertently learn scanner specific, site specific, or individual level cues rather than disease related features, resulting in misleadingly optimistic evaluation outcomes.

Our study contributes to this discussion by explicitly evaluating both individual architectures and an ensemble under identical subject level evaluation conditions. This allows us to assess whether increased model complexity provides any generalization benefit in an extreme low sample regime, or whether simpler architectures are more reliable.

## 2.3 Addressing data scarcity in machine learning

Beyond careful evaluation, a growing body of work has explored strategies for improving learning in data limited settings. Approaches include incorporating prior knowledge into learning systems, generating synthetic data, and designing architectures that are robust to small sample sizes. For example, von Rueden et al. (2022) propose informed machine learning approaches that integrate structured prior knowledge to improve generalization when labelled data are scarce.

In this work, we adopt a deliberately conservative approach. Rather than introducing complex augmentation or generative pipelines, we rely on transfer learning from ImageNet and focus on how architecture choice and evaluation protocol influence performance under severe data constraints. By selecting an efficient model with limited capacity and enforcing strict subject level data separation, we demonstrate that credible and reproducible results are achievable even with very small datasets. This perspective aligns with broader efforts in trustworthy machine learning, where the emphasis is placed not on maximizing headline accuracy, but on ensuring robustness, transparency, and evaluation integrity in settings where data are inherently limited.

## 3. Dataset and Preprocessing
### 3.1 Data source

Resting state functional MRI data were obtained from the Parkinsons Progression Markers Initiative, a large, open access, multi-site research cohort that provides longitudinal imaging and clinical data for Parkinsons disease research (Jennings et al., 2014). PPMI includes healthy controls, early stage Parkinsons patients, and individuals at increased risk for Parkinsons disease. In this study, we focus specifically on the prodromal Parkinsons subgroup, defined by the presence of REM sleep behaviour disorder and related prodromal features in the absence of a clinical Parkinsons diagnosis.

We selected fMRI scans from 20 prodromal subjects, primarily idiopathic REM sleep behaviour disorder cases, and 20 healthy control subjects matched by age and demographic characteristics. All imaging data were acquired under standardized PPMI protocols and accessed through the official data release following institutional approval. For each subject, the resting state fMRI scan consists of a four-dimensional volume with three spatial dimensions and one temporal dimension, acquired while the subject was at rest.

## 3.2 Data representation and slice extraction

The fMRI data were provided in NIfTI format as four-dimensional time series volumes. No additional spatial preprocessing, such as registration to a common template or spatial smoothing, was applied beyond the preprocessing performed by PPMI. Instead, the focus of this work is on downstream data representation and evaluation strategy rather than on optimizing neuroimaging preprocessing pipelines.

To enable the use of standard two-dimensional convolutional neural network architectures, each four-dimensional fMRI volume was converted into a collection of two-dimensional axial slice images. For each subject, axial slices were extracted at each time frame, producing a large set of two-dimensional images that were all assigned the class label of the corresponding subject. This conversion results in several thousand slice images in total. However, these images should not be interpreted as independent samples, as they are highly correlated and originate from only 40 unique individuals.

All slice images underwent minimal intensity normalization and were resized to match the expected input resolution of each network architecture. Images were resized to 224 by 224 pixels for VGG19 and MobileNet V1 (Simonyan and Zisserman, 2014; Howard et al., 2017), and to 299 by 299 pixels for Inception based models (Szegedy et al., 2016; Szegedy et al., 2017). No data augmentation techniques, such as rotations or flips, were applied. This choice was made deliberately to avoid introducing additional correlations between training and test samples, and to ensure that any observed performance differences are attributable to model behaviour and evaluation strategy rather than augmentation effects.

## 3.3 Effective sample size considerations

After slice extraction, the dataset contains on the order of several thousand two dimensional images distributed across two classes. Despite this apparent increase in data volume, the effective sample size in terms of independent observations remains limited to 40 subjects. This distinction between the number of slices and the number of independent subjects is critical for evaluation and model assessment. It directly motivates the subject level data partitioning strategy adopted in this study, which is described in the following section.

## 4. Methods

### 4.1 Overview

The experimental design is structured to examine how evaluation strategy and model capacity influence performance when training deep learning models under extreme data scarcity. Rather than optimizing a single architecture, the goal is to systematically compare convolutional neural networks of varying complexity under different data partitioning schemes, and to quantify the impact of information leakage on apparent generalization.

The pipeline begins with data preparation, where resting state fMRI scans from 20 prodromal Parkinsons subjects and 20 healthy controls are converted from four dimensional volumes into two-dimensional slice images. Each slice inherits the class label of the corresponding subject. These slices are then aggregated into a single dataset, while preserving subject identifiers to enable flexible data partitioning at either the image or subject level.

Two types of dataset splits are considered. In the image level split, slices are randomly assigned to training, validation, and test sets without regard to subject identity. In the subject level split, all slices from a given subject are assigned exclusively to a single partition. Multiple segmentation scenarios are defined to systematically compare these strategies, as detailed in the following section.

A range of convolutional neural network architectures is evaluated, spanning both high capacity and lightweight models. All networks are initialized with ImageNet pretrained weights to mitigate the effects of limited training data. The evaluated architectures include VGG19, Inception V3, Inception ResNet V2, and MobileNet V1, along with an ensemble constructed from two individual models. This selection enables an analysis of how model complexity interacts with data scarcity and evaluation strategy.

For each architecture and each data split configuration, models are fine-tuned on the training set while monitoring performance on a validation set for model selection. Final evaluation is performed on a held-out test set that remains unseen during training and tuning. Model performance is assessed using accuracy, precision, recall, and F1 score for each class, and confusion matrices are examined to identify systematic error patterns.

The results are analysed with a focus on generalization behaviour, performance gaps between image level and subject level evaluation, and indicators of overfitting or information leakage. By repeating experiments across multiple architectures and splitting strategies, the pipeline isolates the effects of data partitioning and model capacity on reported performance.

Figure 1 illustrates the overall workflow, from fMRI volume conversion through model training and evaluation under different data splits. This controlled experimental setup allows a clear assessment of how evaluation rigor and architectural efficiency shape outcomes in extreme low data learning scenarios.

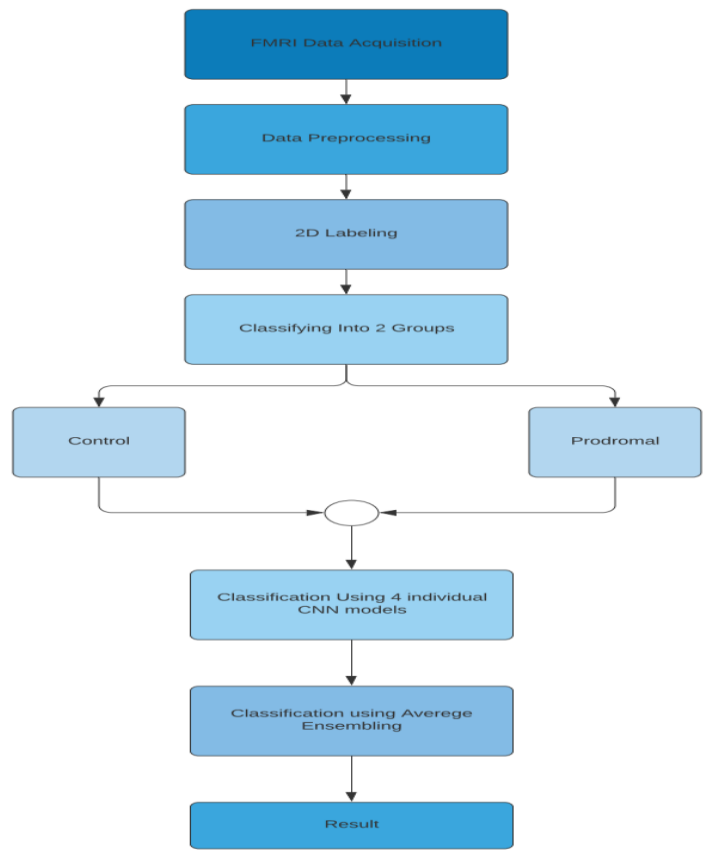

*Figure 1: Schematic of methodology*

### 4.2 Data Splitting Strategies

To study the effect of data partitioning on reported performance, we define three distinct data splitting strategies, referred to as Segmentation 1, Segmentation 2, and Segmentation 3. These strategies are designed to contrast naive image level evaluation with stricter subject level evaluation, and to illustrate how performance estimates can vary when the number of independent test subjects is very small.

**Segmentation 1: Image level split (naive).**
In the first strategy, all two-dimensional slice images are randomly divided into training, validation, and test sets using a 70 percent, 15 percent, and 15 percent split, without considering subject identity. As a result, slices originating from the same subject may appear in multiple partitions. This approach is common in generic computer vision experiments, as it maximizes the number of training samples. However, in the context of medical imaging, it violates subject independence and introduces substantial information leakage. Under this setting, models may rely on subject specific anatomy or scanner related characteristics rather than disease relevant patterns, leading to overly optimistic performance estimates.

**Segmentation 2: Subject level split (single draw).**
In the second strategy, data are partitioned at the subject level. Subjects are randomly assigned to training, validation, and test sets, with 32 subjects (16 prodromal and 16 controls) used for training, 4 subjects (2 per class) for validation, and 4 subjects (2 per class) for testing. All slices from a given subject are confined to a single partition, ensuring that no subject appears in more than one set. This strategy substantially reduces information leakage and provides a more realistic estimate of generalization to unseen individuals. However, because the test set contains only four subjects, performance estimates are inherently noisy and sensitive to which subjects are selected. We use this split as a baseline subject level evaluation.

**Segmentation 3: Subject level split (best case).**
To assess variability in subject level performance, we repeat the random subject level splitting procedure multiple times and identify a split in which test performance is maximized. This scenario represents an optimistic upper bound under subject level evaluation, corresponding to a favorable division of subjects. Importantly, this segmentation is not used for model selection or claims of generalization. Instead, it serves an analytical purpose, illustrating how much reported performance can fluctuate depending on which subjects happen to be included in the test set. In practical settings, such variability would be addressed through repeated cross validation or multiple random splits.

Together, these three strategies allow a direct comparison between naive image level evaluation, a realistic subject level split, and an optimistic subject level scenario. The contrast between Segmentation 1 and Segmentation 2 highlights the magnitude of performance inflation caused by information leakage.

While Segmentation 2 and Segmentation 3 enforce strict subject-level separation between training, validation, and test sets, they nevertheless rely on a single train/validation/test partition for each scenario. Given the total cohort size of 40 subjects, this design implies that performance estimates for subject-level splits are subject to substantial variance and depend on the particular choice of held-out subjects. More exhaustive evaluation protocols, such as leave-one-subject-out cross-validation or repeated random subject-level splits with confidence intervals, were not feasible in this study due to computational and logistical constraints at the time of experimentation. We therefore interpret all subject-level accuracy values as indicative rather than definitive, and focus our conclusions on qualitative patterns (e.g., the relative robustness of lightweight versus high-capacity architectures) rather than precise numerical performance.

### 4.3 CNN Architectures and Training

To examine the relationship between model capacity and generalization under extreme data scarcity, we evaluate a set of convolutional neural network architectures spanning a wide range of complexity. All models are initialized with ImageNet pretrained weights, as the available dataset is far too small to support training from scratch.

The first architecture is VGG1**9** (Simonyan and Zisserman, 2014), a 19-layer convolutional network with a straightforward sequential design consisting of stacked convolution and pooling layers followed by fully connected layers. VGG19 has a very large number of parameters, primarily due to its dense layers, and represents a high-capacity baseline that is prone to overfitting in low data regimes.

The second architecture is Inception V3 (Szegedy et al., 2016), which introduces Inception modules that apply parallel convolutions at multiple scales. Although structurally more complex than VGG,

Inception V3 achieves parameter efficiency through its modular design and requires higher resolution inputs. This model represents a modern, highly optimized architecture designed to balance expressiveness and efficiency.

The third architecture is Inception ResNet V2 (Szegedy et al., 2017), which combines Inception modules with residual connections, resulting in a very deep and computationally intensive network. With substantially higher capacity, this model serves as an upper bound in terms of architectural complexity and allows us to assess whether extreme depth provides any benefit when training data are severely limited.

The final architecture is MobileNet V1 (Howard et al., 2017), a lightweight convolutional network built around depth wise separable convolutions. MobileNet V1 has a substantially smaller parameter count than the other models and is designed for efficiency and implicit regularization. Given its reduced capacity, we hypothesize that MobileNet may generalize more reliably than deeper architectures in this extreme low data setting.

All models are implemented using TensorFlow and Keras and are fine-tuned end to end, with all layers updated during training. A global average pooling layer followed by a single fully connected sigmoid output layer is appended to each network for binary classification. Training is performed using the Adam optimizer with a learning rate of 1e minus 4, a batch size of 16, and a maximum of 20 epochs. Early stopping based on validation loss is applied to mitigate overfitting. Given the near balance between prodromal and control subjects, class weighting is not used. Aside from minimal preprocessing described earlier, no aggressive data augmentation is applied, ensuring that observed effects are attributable to evaluation strategy and model capacity rather than artificial data expansion.

### 4.4 Ensembling

In addition to evaluating individual architectures, we examine whether a simple ensemble can improve generalization in this extreme low data setting. Ensembles are often effective when constituent models learn complementary representations, but in small sample regimes they can also amplify overfitting if all models capture similar spurious patterns. To explore this trade off, we construct a minimal ensemble using two architectures with contrasting capacity profiles: Inception ResNet V2, representing a high-capacity model, and MobileNet V1, representing a lightweight and regularized alternative.

Each network is trained independently using the same training and validation protocol described earlier. After fine tuning, predictions are generated separately for each model. The ensemble output is obtained through late fusion by averaging the predicted posterior probabilities of the positive class. For a given input slice x, if $p\_IRv2(y = 1 \mid x)$ and $p\_MobileNet(y = 1 \mid x)$ denote the predicted probabilities from Inception ResNet V2 and MobileNet V1 respectively, the ensemble prediction is computed as the arithmetic mean of the two probabilities.

No additional training or weighting is applied to the ensemble beyond the training of the individual networks. This simple fusion strategy allows us to assess whether combining models with substantially different capacity leads to more stable predictions, or whether limited data prevent the ensemble from offering meaningful gains. The ensemble is evaluated under the same data splitting strategies as the individual models to ensure a fair comparison.

## 4.5 Evaluation Metrics

Model performance is evaluated on held out test data using a set of standard classification metrics. Accuracy is reported as the primary summary measure, defined as the proportion of correctly classified samples. While accuracy provides an intuitive overview, it can be misleading in the presence of class imbalance or asymmetric error behaviour, and is therefore interpreted alongside additional metrics.

Precision, recall, and F1 score are computed separately for the prodromal and control classes. Precision for the prodromal class measures the proportion of samples predicted as prodromal that are truly prodromal, while recall reflects the proportion of prodromal samples correctly identified. The F1 score provides a balanced summary of precision and recall. For the control class, recall corresponds to specificity. Reporting class wise metrics allows us to identify whether a model favors one class over the other.

Confusion matrices are presented to visualize the distribution of true versus predicted labels. Where appropriate, confusion matrices are normalized by the true class to facilitate interpretation as per class recall. These visualizations help reveal systematic error patterns that may not be apparent from aggregate metrics alone.

For selected experiments, training and validation curves are also examined to assess convergence behaviour and detect overfitting. A widening gap between training and validation performance is taken as an indication of limited generalization.

Because subject level splits involve very few test subjects, results are sometimes additionally summarized at the subject level by assigning a subject level prediction based on the majority vote of that subject's slices. However, slice level metrics remain the primary evaluation unit for consistency across experiments. Where subject level aggregation is used, this is stated explicitly to avoid ambiguity.

## 5. Results

This section reports results under each data splitting strategy, with an emphasis on how evaluation protocol and model capacity affect apparent performance. Unless stated otherwise, all reported metrics correspond to the held-out test set for the respective segmentation.

## 5.1 Segmentation 1: Image level split

When data are split at the image level using a random 70 percent training, 15 percent validation, and 15 percent test partition, all evaluated models achieve extremely high performance. Test accuracy exceeds 99 percent for every architecture, with several models approaching near perfect classification. VGG19, Inception V3, and Inception ResNet V2 reach approximately 99.9 percent test accuracy, while MobileNet V1 and the ensemble also achieve test accuracy close to 99 percent. Precision, recall, and F1 scores for both classes are similarly high, typically in the range of 0.99 to 1.00.

Confusion matrices for this scenario, shown in Figure 2, are nearly identity matrices, with only a small number of misclassified slices out of several thousand test samples. From a purely numerical standpoint, these results would suggest that the classification task has been effectively solved. However, such uniform and near perfect performance across architectures of very different capacity is implausible for prodromal Parkinsons detection and warrants scrutiny.

Inspection of the training dynamics further highlights this issue. Training loss for all models decreases rapidly, often reaching near zero within a few epochs, while validation loss remains comparably low. The absence of a noticeable generalization gap, particularly in a medical imaging task with limited subject diversity, is a strong indication that the evaluation protocol is flawed. Under image level splitting, slices from the same subject appear in both training and test sets, allowing models to exploit subject specific anatomical features or scanner related characteristics rather than learning disease relevant patterns.

As a result, the reported test performance in this setting reflects severe information leakage rather than true generalization. The models effectively memorize subject identity and associated labels, making test predictions trivial once a subject has been seen during training. This explains both the rapid convergence and the near perfect confusion matrices observed in this segmentation.

In summary, the image level split produces highly optimistic and misleading performance estimates. While the numerical results are striking, they do not provide a credible assessment of a model's ability to generalize to unseen individuals. These findings underscore the necessity of subject level evaluation, which we examine next, to obtain a more realistic measure of performance under genuine generalization conditions.

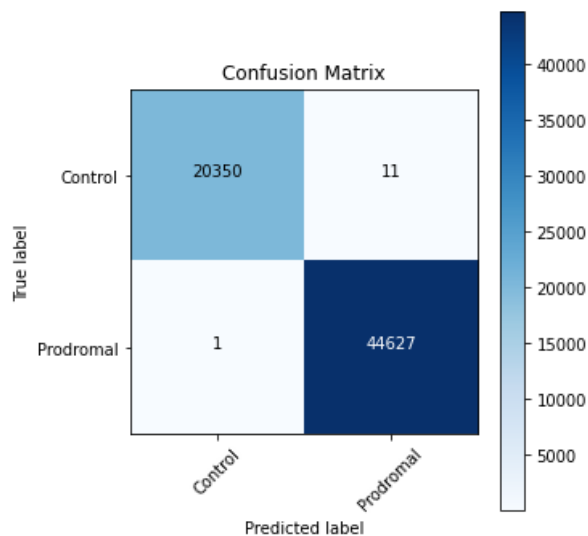

*Figure 2:vgg19 confusion matrix*

## 5.2 Segmentation 2: Subject-Level Split (Single Partition)

When subject identity is strictly enforced across training, validation, and test sets, performance drops sharply to more realistic levels. Under this single random subject level split, none of the evaluated models approach the near perfect results observed with image level splitting. Table 1 summarizes validation and test accuracy for each architecture under this setting.

| Model | Val. Accuracy | Test Accuracy |
|---|---|---|
| VGG19 | 75.40% | 58.36% |

| Model | Val. Accuracy | Test Accuracy |
|---|---|---|
| Inception-V3 | 78.64% | 62.55% |
| Inception-ResNet-V2 | 81.89% | 63.00% |
| **MobileNet-V1** | **83.68%** | **67.20%** |
| Ensemble (IRV2+Mobile) | 77.79% | 63.32% |

Table 1: Summary of validation and test accuracy for Seg 2

Across all models, test accuracy falls in the range of approximately 58 to 67 percent. VGG19 performs the worst, achieving 58.36 percent test accuracy despite substantially higher validation accuracy. Inception V3 and Inception ResNet V2 achieve modest improvements, with test accuracies of 62.55 percent and 63.00 percent respectively. MobileNet V1 yields the strongest performance under this split, achieving a test accuracy of 67.20 percent. The ensemble model does not outperform the best individual architecture, instead achieving performance comparable to Inception ResNet V2.

The substantial gap between validation and test accuracy across all models highlights the high variance inherent in this evaluation setting. With only four subjects in the test set, performance estimates are highly sensitive to which individuals are selected, and even small subject specific effects can lead to large swings in reported accuracy. These results underscore the difficulty of assessing generalization reliably when the number of independent test subjects is extremely limited.

Several observations emerge from this comparison. First, overall test accuracy in the 60 to 67 percent range is consistent with the difficulty of the task, which involves distinguishing subtle and heterogeneous functional patterns associated with prodromal Parkinsons disease. Performance only modestly above chance reflects both the limited statistical power of a 20 versus 20 subject dataset and the intrinsic challenge of the classification problem.

Second, model capacity appears to play an important role. MobileNet V1 consistently outperforms deeper architectures despite having significantly fewer parameters. In contrast, VGG19 performs poorly, likely due to overfitting driven by its large number of parameters and fully connected layers. Inception ResNet V2, although substantially deeper and more expressive, does not translate this additional capacity into improved generalization. These findings suggest that lower capacity models may be better suited to extreme low data regimes.

The ensemble model does not provide a performance gain over the strongest single model. Combining MobileNet V1 with Inception ResNet V2 appears to dilute rather than enhance predictive performance, suggesting that both networks may be learning similar patterns under data constraints, limiting the benefits of ensembling.

Most architectures achieve relatively high sensitivity for the prodromal class, correctly identifying a large proportion of prodromal slices, but struggle to accurately classify control slices. In several cases, prodromal precision and recall are substantially higher than those for the control class, indicating a tendency to overpredict the prodromal label.

This bias may arise from several factors. Subtle class imbalance at the slice level, heterogeneity within the control group, and the aggregation of slice level errors across subjects can all contribute to skewed predictions. In addition, prodromal subjects in this cohort share more consistent clinical characteristics, such as REM sleep behaviour disorder, which may lead models to learn more coherent patterns for that class. These effects combine to favour sensitivity at the expense of specificity.

Overall, Segmentation 2 provides a realistic assessment of model performance under genuine generalization conditions. The best performing model, MobileNet V1, achieves approximately 67 percent accuracy on unseen subjects, indicating moderate predictive ability. Crucially, this evaluation makes clear that the near perfect performance observed under image level splitting is an artifact of information leakage rather than a reflection of true model capability. In the following section, we examine how performance varies across different subject level splits to further characterize this variability.

### 5.3 Segmentation 3: Subject-Level Split (Best-Case Scenario)

To characterize the variability of subject-level performance under extreme data scarcity, we analyze an alternative subject-level partition in which test accuracy is maximized. This scenario represents a favourable division of subjects rather than an independent evaluation and is included solely to illustrate how sensitive reported performance can be when the number of test subjects is very small.

Under this best-case split, all models exhibit higher test accuracy compared to the single-draw subject-level evaluation. MobileNet V1 achieves a test accuracy of 81.22 percent, improving substantially over its performance in the initial subject-level split. Inception ResNet V2 also shows improved generalization under this partition, reaching 75.30 percent test accuracy. Other architectures, including Inception V3 and VGG19, likewise achieve higher accuracy in this setting, though their performance remains below that of MobileNet V1.

The observed improvement highlights the extent to which subject-level performance depends on the specific composition of the test set. With only four test subjects, even a single subject swap between training and test sets can lead to large changes in reported accuracy. Consequently, results obtained from such favourable splits should not be interpreted as evidence of robust generalization, but rather as an optimistic upper bound under particularly accommodating conditions.

Error analysis under this split indicates more balanced classification behaviour than in the initial subject-level evaluation. MobileNet V1 shows improved discrimination between prodromal and control samples, with reduced bias toward predicting the prodromal class. Despite this improvement, misclassifications remain non-negligible, and the model does not approach perfect performance even under these favourable conditions.

Across repeated subject-level splits, MobileNet V1 exhibits test accuracy spanning approximately the high 50s to just over 80 percent, with a mean in the mid-60s. Inception ResNet V2 displays a similar but slightly lower range, while VGG19 and Inception V3 show greater variability and lower average performance. These results reinforce two key observations: first, lightweight architectures such as MobileNet V1 tend to generalize more reliably in extreme low-data regimes; second, performance variance is substantial, and single-split results can be highly misleading.

Overall, the contrast between image-level evaluation, single subject-level evaluation, and best-case subject-level evaluation underscores the importance of reporting performance variability rather than selectively highlighting favourable outcomes. In practical applications, reliable estimation of generalization performance in such settings requires repeated subject-level splits or cross-validation to capture uncertainty and avoid overoptimistic conclusions.

### 6. Discussion

The experiments presented in this work highlight several important considerations for machine learning applied to small-scale biomedical datasets. In this section, we discuss the implications of our findings with respect to evaluation methodology, model selection under data scarcity, limitations of the current study, and broader lessons for trustworthy machine learning in medical imaging.

## 6.1 Impact of Evaluation Strategy and Data Leakage

The most striking result of this study is the large discrepancy between performance obtained under naive image-level evaluation and that obtained under subject-level evaluation. While image-level splitting yields near-perfect accuracy, enforcing subject-level separation leads to substantially lower and more plausible performance. This contrast underscores the critical importance of aligning evaluation strategy with the underlying structure of the data.

When slices from the same subject appear in both training and test sets, models can exploit subject-specific anatomical or acquisition-related features rather than learning disease-relevant patterns. In such cases, high test accuracy reflects memorization rather than generalization. Our results provide a concrete empirical demonstration of this effect: under image-level splitting, all architectures achieve uniformly high accuracy despite large differences in model capacity, whereas under subject-level splitting, performance drops sharply and meaningful differences between models emerge.

Equally important is the degree of variability observed across different subject-level splits. For MobileNet V1, test accuracy ranges from approximately 67 percent to over 80 percent depending on which four subjects are assigned to the test set. This variability highlights the inherent instability of single-split evaluation when the number of independent test subjects is extremely small. The inclusion or exclusion of a single atypical subject can shift reported performance by more than ten percentage points, making point estimates unreliable.

These observations reinforce the need for evaluation protocols that explicitly account for uncertainty. In small-sample settings, repeated subject-level splits or cross-validation strategies, such as leave-one-subject-out evaluation, provide a more informative assessment by capturing the distribution of possible outcomes rather than a single optimistic result. While full cross-validation was not implemented in the present study due to practical constraints, the multiple segmentation analysis already illustrates the magnitude of variance that such approaches are designed to capture.

The broader implication is that trustworthy machine learning depends as much on evaluation design as on model architecture. For subject-based inference tasks, evaluation must be subject-based. Failure to enforce this principle can lead to severe information leakage and inflated performance claims that do not translate to real-world deployment. Our findings add to a growing body of evidence showing that rigorous data partitioning is essential in medical imaging studies, particularly when datasets are small and highly correlated.

From a practical perspective, this study serves as a cautionary example. Had we relied on image-level results alone, the conclusion would have been that prodromal Parkinsons disease detection from fMRI is a nearly solved problem. Subject-level evaluation reveals a far more nuanced and realistic picture. We therefore advocate that medical imaging studies explicitly verify subject independence throughout the entire pipeline, including data augmentation and preprocessing, and prioritize conservative evaluation strategies when uncertainty exists.

## 6.2 Model Complexity and Generalization in Low-Data Regimes

The results of this study offer clear insights into model selection when training data are extremely limited. Across subject-level evaluations, the lightweight MobileNet V1 consistently outperforms substantially deeper architectures. Under the single subject-level split, MobileNet achieves the highest test accuracy among all models, and under the most favourable split it reaches over 80 percent accuracy, exceeding the performance of much larger networks such as Inception ResNet V2. These findings suggest that, when training data consist of only a few dozen subjects, reduced model capacity can be an advantage rather than a limitation.

Large architectures such as Inception ResNet V2 offer high representational power, but this capacity comes at the cost of many trainable parameters. In extreme low-data settings, such models may struggle to tune their parameters effectively and are more prone to overfitting. The relatively modest gains observed for Inception ResNet V2, compared to its much higher complexity, indicate that additional depth and architectural sophistication do not necessarily translate into improved generalization when the available data cannot support them.

MobileNet V1, by contrast, is explicitly designed for efficiency. Its use of depth wise separable convolutions, reduced parameter count, and streamlined architecture introduce an implicit form of regularization that appears well suited to data-scarce learning scenarios. Notably, although MobileNet achieves lower accuracy than deeper architectures on large-scale benchmarks such as ImageNet, it generalizes more reliably in this task. This observation is consistent with prior work in medical imaging showing that simpler models can outperform larger ones when training data are limited and heterogeneous.

The poor performance of VGG19 further reinforces this point. Despite its relatively simple structure, VGG19 contains a very large number of parameters due to its fully connected layers. In this setting, the model appears to overfit the training data and fails to generalize to unseen subjects. Inception ResNet V2 performs better than VGG19 but does not surpass MobileNet, suggesting that architectural refinements such as residual connections and extreme depth are insufficient to overcome data limitations.

The ensemble of MobileNet V1 and Inception ResNet V2 does not improve upon the best individual model. Instead, its performance lies between that of the two constituent networks. This outcome suggests that, under severe data constraints, ensemble methods may offer limited benefit, particularly when base models learn similar patterns or share the same sources of error. In such cases, ensembling effectively increases model capacity without introducing genuinely complementary representations, potentially reintroducing overfitting rather than mitigating it.

Overall, these findings argue in favour of prioritizing architectural simplicity and efficiency when learning from extremely small datasets. While there is a strong tendency in modern deep learning to favour increasingly large and complex models, such approaches may be counterproductive when the data regime does not support them. Lightweight architectures provide not only improved generalization in this setting, but also faster training, lower computational cost, and greater experimental flexibility. As additional data become available, model capacity can be scaled accordingly, but in the absence of sufficient data, efficient models represent a more reliable and principled choice.

### 6.3 Limitations and Future Work

The most fundamental limitation of this study is the very small cohort size: only 40 subjects (20 prodromal, 20 controls). This extreme data scarcity is central to the motivation of the work, but it also constrains the strength of any quantitative claims: statistical power is limited, and performance estimates are highly sensitive to which subjects are assigned to training, validation, and test sets. Although we explore multiple subject-level splits and explicitly contrast single-draw and "best-case" scenarios, we do not perform comprehensive leave-one-subject-out cross-validation or report confidence intervals. Consequently, the absolute accuracy values reported here should not be interpreted as precise estimates of real-world performance; instead, they serve to illustrate the magnitude of performance inflation under image-level splitting and the relative behaviour of architectures with different capacities under consistent subject-level evaluation.

A second limitation is the absence of external validation on independent cohorts. All experiments are conducted on a single open-access dataset, and no additional prodromal Parkinson's fMRI collections were available at the time of this work. This means that we cannot empirically assess robustness to changes in acquisition protocols, sites, or population demographics. Nonetheless, the methodological conclusions—regarding the dangers of image-level splitting and the benefits of lightweight architectures under strict subject-level evaluation—are rooted in general properties of correlated, hierarchical data and are therefore expected to transfer qualitatively to other small-sample settings, even if the exact numerical performance does not.

The methodological choice to convert four-dimensional fMRI data into two-dimensional slice images also introduces limitations. While this approach enables the use of standard convolutional architectures and simplifies the experimental setup, it discards potentially informative spatial and temporal structure present in the full fMRI volumes. Resting-state functional connectivity patterns are inherently spatiotemporal, and approaches that operate directly on three-dimensional volumes or model temporal dynamics may capture richer disease-related signals. However, training such models reliably with only dozens of subjects remains challenging due to increased dimensionality and parameter count. Promising future directions include hybrid approaches that combine dimensionality reduction, such as functional connectivity features or autoencoder-based representations, with lightweight classifiers, as well as transfer learning from large-scale neuroimaging datasets.

Another important limitation concerns model interpretability. This study focuses primarily on evaluation rigor and generalization behaviour and does not include an in-depth analysis of what features or regions the models rely on when making predictions. In medical contexts, interpretability is essential for building trust and for identifying potential failure modes, particularly when data leakage or spurious correlations are a concern. Techniques such as saliency mapping or gradient-based attribution could be applied in future work to assess whether the models attend to meaningful neurobiological patterns or to subject-specific artifacts.

Finally, while prodromal Parkinson's detection provides a clinically meaningful testbed, this work should not be interpreted as proposing a ready-to-deploy diagnostic system. An accuracy of around 80% in a favourable split, obtained without external validation or prospective evaluation, falls well short of clinical requirements, particularly given that the prodromal cohort is defined largely by REM sleep behaviour disorder, a strong predictor. For real-world use, any imaging-based model would need to demonstrate clear added value over simple clinical predictors and be validated across multiple cohorts and acquisition conditions. Our focus here is therefore on evaluation integrity and model selection under extreme data scarcity, rather than on clinical translation.

Despite these limitations, the methodological insights from this study remain valuable. The results demonstrate how easily misleading conclusions can arise from inappropriate evaluation in small-sample settings, and how enforcing strict subject-level validation can guide more principled modelling choices. The finding that a lightweight architecture can generalize better than deeper models under extreme data scarcity is encouraging and suggests that meaningful signal exists even in very limited datasets. With additional data, improved validation protocols, and techniques that incorporate prior knowledge, future work may build on these insights to develop more robust and reliable models for early disease detection.

## 6.4 Toward Trustworthy Machine Learning in Low-Data Medical Settings

The experience gained through this study reinforces several core principles for developing trustworthy machine learning systems in medical contexts, particularly when data are scarce. Foremost among these is transparency in evaluation. By explicitly demonstrating how near-perfect performance can arise from flawed evaluation protocols, and by contrasting these results with more realistic subject-level estimates, we aim to clarify which findings are meaningful and which are artifacts. Openly reporting such discrepancies, even when they reveal overly optimistic initial results, is essential for maintaining scientific credibility and for enabling reproducibility.

A second key principal concerns model simplicity. In low-data regimes, simpler and more efficient models are often not only easier to train and validate, but also more robust. Lightweight architectures such as MobileNet V1 impose implicit regularization through reduced capacity, making them less prone to memorization and easier to analyze. In safety-critical applications, a model that is modest in complexity and well understood may be preferable to a more powerful but brittle alternative, even if nominal accuracy is slightly lower. In this study, the most efficient model also proved to be the most reliable under proper evaluation, reinforcing the case for prioritizing efficiency over architectural sophistication when data are limited.

Finally, rigorous validation remains a prerequisite for any claims of real-world applicability. While the subject-level evaluation adopted here represents a substantial improvement over naive slice-based splitting, further steps are necessary to approach deployment-level confidence. Independent test sets, cross-study validation, and prospective evaluation are particularly important in medical settings where subtle biases or acquisition artifacts can strongly influence outcomes. In addition, systematic analysis of failure cases may provide valuable insights into model limitations and data quality issues. Understanding why specific subjects are consistently misclassified could inform improvements in both data collection and modelling strategies.

Taken together, this work demonstrates how a biomedical application can be reframed as a machine learning–centric investigation focused on evaluation integrity, generalization under data scarcity, and responsible model selection. The results show that meaningful learning is possible even with extremely limited data, if models are chosen appropriately and evaluation protocols are carefully designed. More broadly, this study contributes to ongoing efforts within the machine learning community to ensure that reported results are not only impressive, but also reliable, reproducible, and representative of real-world performance. We hope that this work serves as a practical reference for researchers working with small medical datasets and encourages the adoption of conservative, transparent methodologies in similarly constrained settings.

## X Reproducibility and evaluation practices in low-data settings

Although the experiments in this study were conducted in a constrained computational environment and we do not release code or pre-processed data; we aim to make the evaluation protocol as transparent and reproducible as possible. We explicitly describe the subject-level partitioning strategies, the network architectures and initialization schemes, and the training hyperparameters (optimizer, learning rate, batch size, and number of epochs). In future work, we advocate moving toward standardized low-data benchmarks that provide fixed subject-level splits, reference implementations, and recommended reporting formats (e.g., cross-validation with confidence intervals and statistical tests), in line with recent proposals for experimental standards in machine learning evaluation. For domains where data cannot be openly shared, releasing code and detailed configuration files for preprocessing and model training remains essential to enable meaningful comparison and replication.

## 7. Conclusion and Future Work

This work presents a methodological case study on prodromal Parkinson's disease detection from resting-state fMRI, framed as a problem of learning under extreme data scarcity. Rather than emphasizing biomedical novelty, the study focuses on challenges that arise when deep learning models are trained and evaluated on highly limited, correlated data. Two central findings emerge: the necessity of subject-level evaluation to obtain meaningful performance estimates, and the effectiveness of lightweight convolutional neural networks in low-data regimes.

Through systematic comparison of data partitioning strategies, we show that naive image-level splitting can produce deceptively high accuracy approaching 100 percent due to severe information leakage. When subject-level separation is enforced, performance drops to a more realistic range of approximately 60 to 80 percent, revealing the true difficulty of the task. These results demonstrate how easily evaluation misuse can lead to misleading conclusions and underscore the importance of aligning validation protocols with the hierarchical structure of the data.

Across all subject-level evaluations, MobileNet V1 consistently outperforms substantially deeper architectures. Despite having far fewer parameters, the lightweight model achieves the strongest generalization performance, including a best-case accuracy of just over 80 percent under favourable subject partitions. In contrast, larger models such as Inception ResNet V2 fail to translate their additional capacity into improved performance. This finding reinforces the principle that, when data are scarce, architectural efficiency and implicit regularization can be more valuable than representational power. At the same time, the small cohort size and lack of exhaustive cross-validation mean that absolute accuracy values—particularly best-case results—should be interpreted as optimistic upper bounds rather than precise estimates of out-of-sample performance.

Beyond the specific application, this study highlights broader lessons for machine learning research in data-limited settings. Careful design choices, including strict data partitioning, conservative model selection, and restrained interpretation of results, enable meaningful learning even when datasets are small. The observed variability across subject-level splits illustrates why single-point performance estimates are insufficient and why uncertainty should be explicitly acknowledged, for example through cross-validation, confidence intervals, or repeated runs.

Several directions for future work follow naturally from this analysis. A priority is to perform subject-level cross-validation, such as leave-one-subject-out evaluation, to provide more stable performance estimates and quantify uncertainty more rigorously. Another direction is to explore models that operate on higher-dimensional representations of fMRI data, including three-dimensional or spatiotemporal approaches, potentially enabled by transfer learning or unsupervised pretraining on larger neuroimaging datasets. Incorporating domain knowledge, such as known functional brain networks or graph-based representations, may further improve robustness in low-data regimes.

From an application perspective, the methodological insights developed here are not specific to prodromal Parkinson's disease. Similar challenges arise in other neurological conditions with small imaging cohorts, including prodromal Alzheimer's disease and rare disorders. Extending this framework to such settings, and to multimodal data that combine imaging with clinical or behavioural information, represents a promising avenue for future research.

In conclusion, this work demonstrates that credible machine learning results are achievable even under extreme data constraints, if evaluation integrity and model simplicity are prioritized. Lightweight architectures combined with rigorous subject-level validation offer a principled foundation for exploratory modelling in small-sample medical studies. More broadly, the lessons presented here contribute to ongoing efforts to promote trustworthy, reproducible machine learning practices in domains where data are limited and the cost of overconfidence is high.

**Acknowledgments**

This work is based on the original undergraduate thesis research by Naimur Rahman and colleagues at Brac University (2017–2021). The authors thank Dr. Muhammad Zavid Parvez and Md. Tanzim Reza (Brac University) for their invaluable supervision of that thesis project. We also acknowledge the Michael J. Fox Foundation and the Parkinson's Progression Markers Initiative (PPMI) for providing the open-access fMRI data used in this study. Data were obtained from the PPMI database (www.ppmi-info.org/data) – PPMI is sponsored and partially funded by the Michael J. Fox Foundation. Finally, we extend our thanks to the co-authors of the original thesis – Farhan Shahriar, Amarttya Prasad Dey, Zarin Tasnim, and Mohammad Zubayer Tanvir – for their contributions to the initial experimental work on which this paper is built.